\documentclass{article}

\usepackage{microtype}
\usepackage{graphicx}
\usepackage{booktabs} 

\usepackage{hyperref}

\usepackage[utf8]{inputenc}

\usepackage{microtype}

\usepackage{inconsolata}

\usepackage{braket}
\usepackage{graphicx}
\usepackage{amsmath}
\usepackage{amssymb}
\usepackage{booktabs}
\usepackage{multirow}
\usepackage{multicol}
\usepackage{tabularx}
\usepackage{arydshln}
\usepackage{makecell}
\usepackage[T1]{fontenc}

\usepackage{cleveref}
\usepackage{mdframed}
\usepackage{xcolor}
\usepackage{listings}

\usepackage{capt-of}
\usepackage{subcaption}

\lstnewenvironment{text-sample}[2]
  {%
   \mdframed[
        backgroundcolor=gray!20,
        skipabove=\topsep
   ]
   \lstset{
     basicstyle=\ttfamily\small, 
     breaklines=true,
     columns=fullflexible,
     escapeinside = ||,
     breakindent=0pt
   }%
  }
  {\endmdframed}

\title{Long-Form Speech Generation with Spoken Language Models}

\usepackage[accepted]{icml2025}

\usepackage{amsmath}
\usepackage{amssymb}
\usepackage{mathtools}
\usepackage{amsthm}
\usepackage{pifont}

\theoremstyle{plain}

\theoremstyle{definition}

\theoremstyle{remark}

\icmltitlerunning{Long-Form Speech Generation with Spoken Language Models}

\begin{document}

\twocolumn[
\icmltitle{Long-Form Speech Generation with Spoken Language Models}



\icmlsetsymbol{equal}{*}

\begin{icmlauthorlist}
\icmlauthor{Se Jin Park}{equal,gdm,ivllab}
\icmlauthor{Julian Salazar}{equal,gdm}
\icmlauthor{Aren Jansen}{gdm}
\icmlauthor{Keisuke Kinoshita}{gdm}
\icmlauthor{Yong Man Ro}{ivllab}
\icmlauthor{RJ Skerry-Ryan}{gdm}
\end{icmlauthorlist}

\icmlaffiliation{gdm}{Google DeepMind.}
\icmlaffiliation{ivllab}{Integrated Vision and Language Lab, KAIST}

\icmlcorrespondingauthor{Se Jin Park}{jinny960812@kaist.ac.kr}
\icmlcorrespondingauthor{Julian Salazar}{julsal@google.com}

\icmlkeywords{Speech Generation, Spoken Language Models, Machine Learning, ICML}

\vskip 0.3in
]



\printAffiliationsAndNotice{\textsuperscript{*}Primary contributors. Se Jin's work was done as a student researcher at Google DeepMind.}  

\begin{abstract}
We consider the generative modeling of speech over multiple minutes, a requirement for long-form multimedia generation and audio-native voice assistants. However, textless spoken language models struggle to generate plausible speech past tens of seconds, due to high temporal resolution of speech tokens causing loss of coherence, architectural issues with long-sequence training or extrapolation, and memory costs at inference time. From these considerations we derive \textbf{SpeechSSM}, the first 
speech language model family to learn from and sample long-form spoken audio (e.g., 16 minutes of read or extemporaneous speech) in a single decoding session without text intermediates. SpeechSSMs leverage recent advances in linear-time sequence modeling to greatly surpass current Transformer spoken LMs in coherence and efficiency on multi-minute generations while still matching them at the utterance level.
As we found current spoken language evaluations uninformative, especially in this new long-form setting, we also introduce: \textbf{LibriSpeech-Long}, a benchmark for long-form speech evaluation; new embedding-based and LLM-judged metrics; and  quality measurements over length and time. Speech samples, the LibriSpeech-Long dataset, and any future code or model releases can be found at \url{https://google.github.io/tacotron/publications/speechssm/}.
\end{abstract}

\section{Introduction}

Generative spoken language models \citep{lakhotia2021generative, dieleman2021-variable, oord2017-vqvae} are autoregressive models of invertible audio representations, enabling the direct learning and generation of intelligible speech and its paralinguistic aspects, such as prosody \citep{kharitonov2022-pgslm} and turn-taking \citep{nguyen2023generative}. These capabilities make speech-native language models (LMs) promising for applications like media understanding and co-creation, audio-native voice assistants, and textless NLP.
However, real-world use-cases of spoken LMs require the ability to both understand and generate long-form speech. For example, voice interactions can last many minutes, requiring a model to maintain a growing conversational history in real time, and expressive media like audiobooks and podcasts can require semantic, paralinguistic, and speaker coherence over a chapter or episode.

\begin{figure}[t]
	\begin{minipage}[b]{\linewidth}
		\centering		
          \centerline{\includegraphics[clip, trim=0 0 0 0, width=\linewidth]{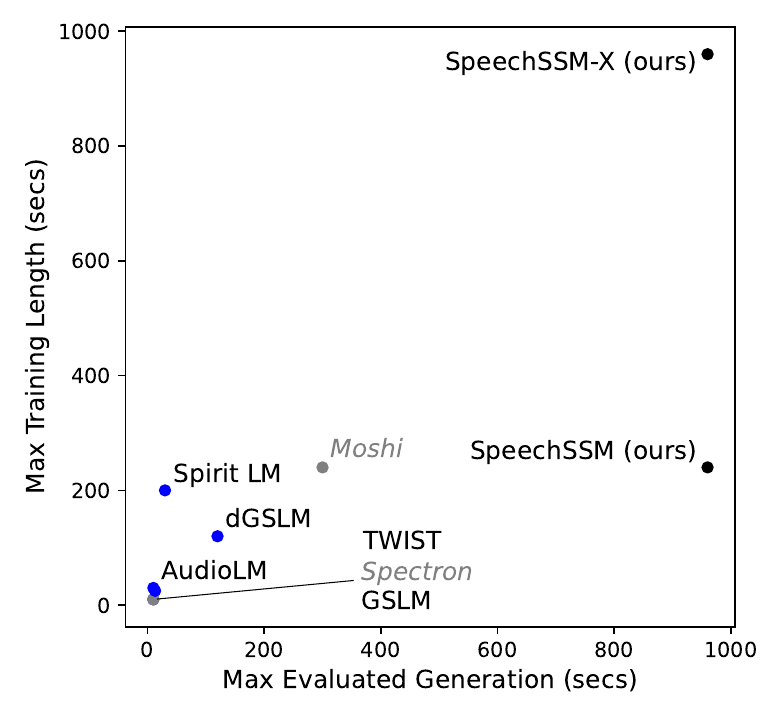}}
	\end{minipage}
    \vspace{-1.0cm}
  \caption{Maximum sequence lengths considered by various spoken LMs. \textit{Italicized} models used text intermediates at generation time. Our models can generate indefinitely due to their constant memory footprint, but we cap our evaluations to 16 minutes.}
  \vspace{-0.5cm}
  \label{fig:prior_work}
\end{figure}

\begin{figure*}[t]
	\begin{minipage}[b]{\linewidth}
		\centering		
            \centerline{\includegraphics[width=0.95\linewidth]{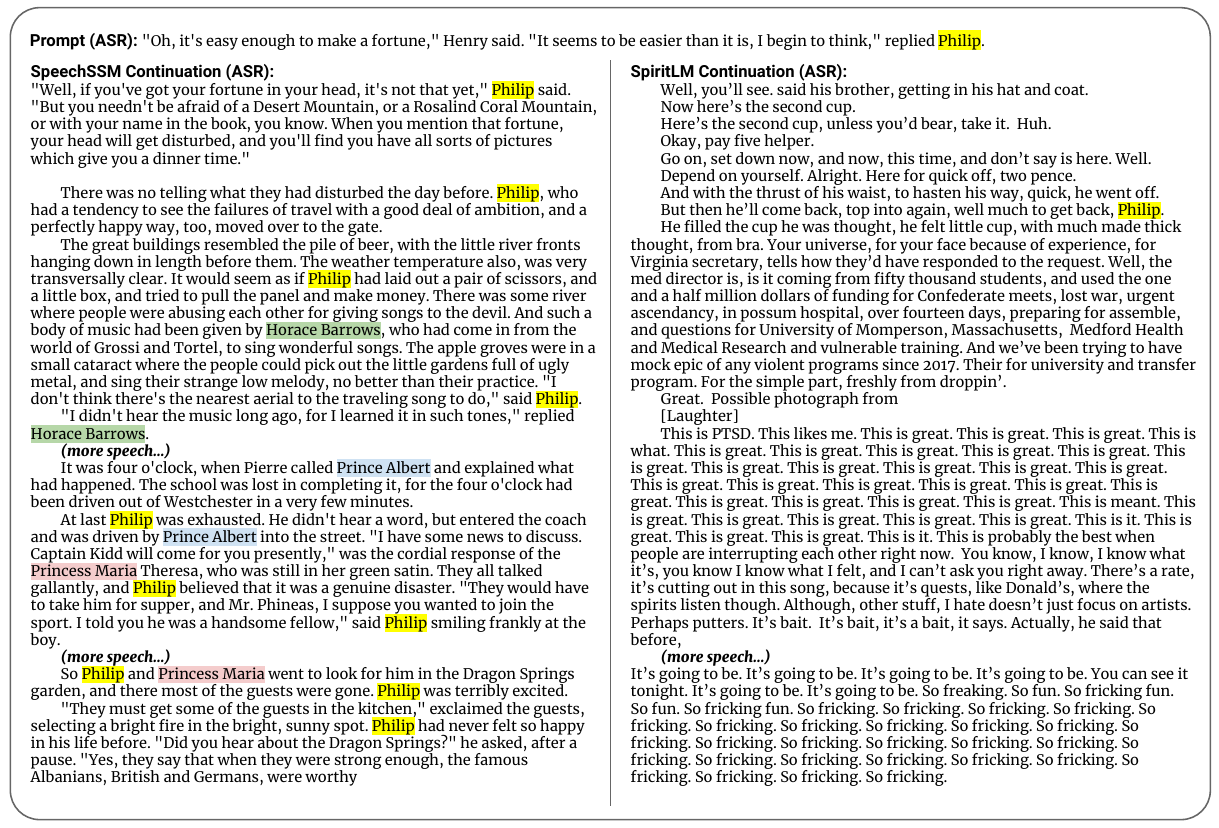}}
	\end{minipage}
    \vspace{-0.7cm}
	\caption{Automated transcripts of 4min speech continuations generated by SpeechSSM-2B (ours) and a Spirit LM Expressive (7B) model \citep{nguyen2024spirit} under slide-and-prompt generation (\Cref{sec:long-form-exps}), extending a 10-second audio-only prompt from our proposed LibriSpeech-Long (test-clean). Aspects like recurring proper nouns show SpeechSSM's relative semantic consistency over time.}
  \vspace{-0.3cm}
  \label{fig:hero-transcript}
\end{figure*}

This presents significant challenges for existing spoken language models, as spoken audio's
textual content is entangled with paralinguistic and acoustic properties that detract from learning invariant representations of meaning. Additionally, current audio representations have high temporal rates, requiring 10+ speech tokens to cover the duration of 1-2 text tokens \citep{hassid2023textually}. Hence, models must disentangle, aggregate, and generate content coherently over longer time horizons. A single-stage Transformer \citep{vaswani2017-transformer} LM is impractical, as its initial cost grows quadratically with prompt length, and its per-step cost grows linearly when decoding. Furthermore, it may also be ineffective, as suggested by Transformer's degraded performance on long-range tasks \citep{tay2021-lra}. Though a few works have improved speech coherence via joint modeling with text (\Cref{sec:related-work}), the challenge of directly modeling long-form speech, particularly generation, remains unstudied by existing spoken LMs (\Cref{fig:prior_work}) and the field overall. Our work proposes and makes initial progress on generative long-form speech:

\textbf{Modeling.} We discuss the design choices required to enable the practical training, generation, and extrapolation to tens of minutes of audio, from tokenization to speaker conditioning to complexity with respect to sequence length. The result is \textbf{SpeechSSM}, a new (textless) spoken language model family (2B, 9B) designed for long-form generation, being the first to model and generate unbounded long-form speech in bounded memory and the first state-space spoken LM. As baselines, we also train spoken Transformer LMs to perform multi-minute generations. Finally, we demonstrate \textbf{SpeechSSM-X}, an extemporaneous variant for naturalistic spontaneous speech.

\textbf{Evaluation.} We observe that existing metrics in speech generation evaluation are noisy and poorly discriminative, and propose the use of reference-based semantic metrics, side-by-side LLM-as-judge, and time-stratified evaluations for speech generation. To scale these to long-form evaluation, we introduce the \textbf{LibriSpeech-Long} benchmark, which reprocesses LibriSpeech's \cite{panayotov2015librispeech} dev and test sets' original chapter-level audio into utterance-aligned spans of up to 4 minutes. This enables much longer prompts and ground truths for reference-based evaluations in tasks like long-form speech continuation, speech recognition, and text to speech.

We find that SpeechSSM matches existing spoken LMs on short generations, while outperforming their sliding window extensions on long generations (e.g., \Cref{fig:hero-transcript}), and that our proposed metrics and benchmark distinctly quantify the quality gaps between past work, our work, and human-level speech generation---enabling future development. We release examples
of read- and extemporaneous-style generations of up to 16 minutes in length,\footnote{\href{https://google.github.io/tacotron/publications/speechssm/}{https://google.github.io/tacotron/publications/speechssm/}}
and we release the LibriSpeech-Long evaluation dataset\footnote{\href{https://github.com/google-deepmind/librispeech-long/}{https://github.com/google-deepmind/librispeech-long/}}
under a CC-BY 4.0 license.

\section{Related Works}
\label{sec:related-work}
\label{sec:related-work-evals}

\textbf{Generating with Spoken LMs.} The family of GSLM models \cite{lakhotia2021generative, kharitonov2022-pgslm} are Transformer decoder LMs trained on discrete units obtained from $k$-means clustering of HuBERT \cite{hsu2021hubert} features and synthesized via unit-to-spectrogram or unit-to-waveform models. This approach gave promising temporal coherence but poor audio quality, and so AudioLM \cite{borsos2023audiolm} proposed separate LMs, one for semantic tokens as before, and two for modeling coarse-to-fine acoustic tokens that are residual codes of a neural audio codec \citep{zeghidour2021soundstream}; this was simplified and made non-autoregressive by \citet{borsos2023soundstorm}. 
TWIST \citep{hassid2023textually} found that text LM initialization improved content-level coherence, atop which VoxtLM \citep{maiti2023voxtlm} and Spirit LM \citep{nguyen2024spirit} found that joint or interleaved training with text gave further improvements.

Beyond the scope of this work, there are audio-text models like SpeechGPT \citep{zhang2023speechgpt} trained for sequence-to-sequence and not generative continuation; there are also dual-channel and text-intermediate models like dGSLM \cite{nguyen2023generative} whose semantic evaluations are <20s, Spectron \citep{nachmani2023spoken} which passes through text, and Moshi \citep{defossez2024moshi} which had few-minute dialogues via time-aligned text.

\textbf{State-Space Models for Long-Form Audio.} State-space models (SSMs; \citealp{gu2021combining}) have become popular among efficient (sub-quadratic) replacements for Transformer-based architectures, giving the first model (S4; \citealp{gu2021efficiently}) to perform all tasks in the Long-Range Arena \citep{tay2021-lra}, outperforming the vanilla Transformer. They utilize constant computation and memory requirements to generate tokens during inference and can be efficiently trained. Recent focus has shifted to hybrid models \cite{glorioso2024zamba, lieber2024jamba, de2024griffin} which integrate state-space layers and variants like linear recurrent units (LRU; \citealp{orvieto2023-lru}) with finite-context self-attention layers.
Recent works have considered SSMs in audio, primarily to support long speech \textit{inputs} for text-out tasks like automatic speech recognition (ASR) and summarization. None are spoken LMs for speech continuation, with only one considering (acoustic-level) tokens \citep{gao2024speech}; most works involve spectrogram encoders or outputs \cite{shams2024ssamba, erol2024audio, lin2024audio, miyazaki2024exploring}. Closest in spirit is SaShiMi \citep{goel2022s}, a multi-scale S4 operating directly on waveform samples; though they generated only 1s of speech, this corresponds to a sequence of 16k discretized scalars.

\textbf{Evaluating Spoken LM Generations.} \citet{lakhotia2021generative} was first to evaluate the generations of spoken LMs, proposing ASR as a path to automated text metrics like text perplexity (PPL) and proportion of repeated $k$-grams (auto-BLEU), along with human evaluations of intelligibility and meaningfulness with mean opinion scores (MOS and MMOS respectively). For their spoken LMs, zero-shot (non-generative) metrics based on logprobs of contrastive pairs (sWUGGY and sBLIMP; \citealp{nguyen2020zero}) were predictive of generation performance, though scores varied with token vocabulary size.
However, these initial metrics seem to lack robustness or are saturating with respect to newer spoken LMs. \citet{hassid2023textually} found transcript PPL and auto-BLEU to be noisy, favoring MMOS and expanding zero-shot metrics (sStoryCloze and tStoryCloze). In turn, \citet{defossez2024moshi} found that sWUGGY and sBLIMP scores degraded despite experiential improvement from noise augmentation and instruction-tuning, instead favoring spoken question-answering \citep{nachmani2023spoken} evaluated via ASR.

Closest to our work was the use of LLMs to assign absolute, reference-free scores to assess the instruction-following of turn-based text-and-speech LMs in \citet{zhang2023speechgpt, zhang2024speechagents}.
As for observing saturation, \citet{borsos2023audiolm} found that humans could not distinguish between a synthetic 7s continuation versus the real 7s continuation of a 3s prompt on a holistic side-by-side evaluation, suggesting the need for more targeted and longer-form evaluations.

\section{Unbounded Speech Generation}
\label{sec:unbounded-speech-gen}

\begin{figure*}[t]
	\begin{minipage}[b]{\linewidth}
		\centering		
            \centerline{\includegraphics[width=\linewidth, trim={0 0 0 0}, clip]{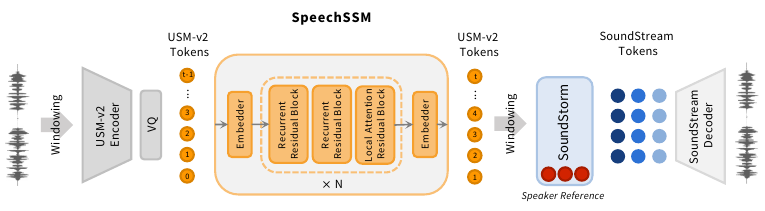}}
	\end{minipage}
    \vspace{-0.7cm}
	\caption{Overview of SpeechSSM. \textit{Left:} A causally-masked hybrid state-space model (Griffin) is trained with an LM objective on semantic tokens (USM-v2) encoded via overlapping fixed-size windows. \textit{Right:} A non-autoregressive synthesizer (SoundStorm) converts overlapping windows of semantic tokens to the acoustic tokens of a neural codec (SoundStream) in a speaker-conditioned manner.}
  \label{fig:architecture}
\end{figure*}

We begin by proposing a set of requirements for a general, unbounded, speech generation system:
\begin{itemize}
    \item \textbf{Constant memory during decoding}, to enable indefinite AR sampling without running out of memory.
    \item \textbf{Infinite context}, so that arbitrarily distant dependencies can be expressed, at least in theory. With the above, this means relevant context must fit in a fixed-size state.
    \item \textbf{Generative length extrapolation}, so that speech quality remains consistent over time, in particular beyond audio durations seen during training.
\end{itemize}
The first leads us to linear-complexity sequence modeling with a fixed-size state. The second leads us to models with aggregation mechanisms such as recurrences or compressive memories. We show that with some care, one can also achieve the third requirement of generative extrapolation.

Finally, there is also a soft requirement for \textit{efficient training}, e.g., train-time dependence on sequence length that is subquadratic, to enable longer sequences and reduce reliance on extrapolation. This favors a parallelizable weight learning scheme, which naturally leads to state-space models (broadly defined, i.e., including linear recurrence models and certain hybrid variants; \citealp{patro2024mamba360, dao2024transformers}) and thus SpeechSSM, a family of hybrid state-space spoken language models for efficient long-form speech generation that fulfills all these desiderata:

\textbf{Architecture.} For our decoder-only hybrid SSM we choose Griffin \cite{de2024griffin}, which interleaves a gated variant of LRUs \cite{orvieto2023-lru} and local (sliding-window) multi-query attention (MQA) blocks in a fixed pattern (two recurrent, one local-MQA; see \Cref{fig:architecture}, left). Local attention efficiently captures recent context, while the states of the gated recurrences transmit information across arbitrary distances. Griffin's performance matched comparable Transformers while greatly improving inference speed and enabling context-side extrapolation at least 4x longer than seen in training. As RoPE \cite{su2024roformer} in the local-MQA blocks still encodes absolute position, we follow recent work on position embeddings (PEs) under causal self-attention (NoPE; \citealp{kazemnejad2024impact}) and remove all explicit PEs from SpeechSSM to promote extrapolation.

\textbf{Initialization.} Inspired by \citet{hassid2023textually}'s success with text-initialized spoken language models (TWIST), we initialize our models with RecurrentGemma-\{2B,9B\} IT \cite{botev2024recurrentgemma}, which are open-weight LMs with the Griffin architecture, trained on 2 trillion text tokens. We discard the pretrained text token embeddings and initialize new ones for our audio token vocabulary.

\textbf{Semantic Tokenizer.} We use the pretrained USM-v2 speech tokenizer \citep{vashishth2024stab, rubenstein2023audiopalm}. Its encoder \citep{zhang2023-usm} is trained with masked language modeling on untranscribed audio and an auxiliary ASR  loss on transcribed audio. Inner representations are vector-quantized into 32k units that serve as fixed-rate (25Hz) pseudo-text for our speech LM. \citet{vashishth2024stab} found that USM-v2 was by far the most speaker-invariant token among common speech tokenizers.

\textbf{Speaker-Prompted Audio Synthesis.} Following \citet{borsos2023audiolm}, we have a second stage that generates low-level acoustic tokens conditioned on semantic tokens. We use a SoundStorm model \cite{borsos2023soundstorm} to non-autoregressively generate SoundStream tokens \cite{zeghidour2021soundstream}, a standard neural audio codec that efficiently reconstructs to high-quality audio. Notably, one can train SoundStorm to support 3s voice prompts (a frozen prefix of semantic and acoustic tokens), such that output acoustic tokens reflect its speaker characteristics.

By choosing a token and model decomposition that isolates speaker characteristics to the acoustic stage, SpeechSSM focuses capacity on modeling semantic coherence along the temporal axis.

\begin{figure}[t]
    \vskip -0.2cm
    \centering
    \begin{subfigure}[t]{0.50\columnwidth}
    \centering
    \includegraphics[width=3.6cm]{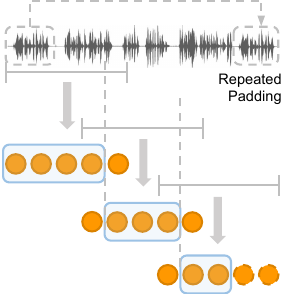}
    \vspace{-0.1cm}
    \caption{Tokenization}
    \label{fig:tokenization}
    \end{subfigure}
    \hfill
    \begin{subfigure}[t]{0.48\columnwidth}
    \centering
    \includegraphics[width=3.2cm]{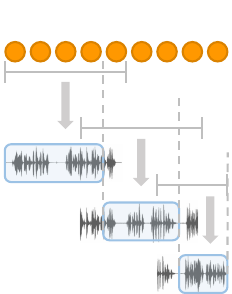}
    \vspace{-0.1cm}
    \caption{Decoding}
    \label{fig:synthesis}
    \end{subfigure}
    \vspace{-0.2cm}
    \caption{Depiction of how input and output windowing work, shown here with 5-token window widths and 2-token overlaps.}
    \vspace{-0.5cm}
\end{figure}

\textbf{Windowed Tokenization and Decoding.} To process long-form speech while bounding the memory of the (non-SSM) semantic tokenizer and acoustic decoder, we divide audio into fixed segments, with overlaps between neighbors. Each window is tokenized independently, then merged into a single stream at each overlap by taking the first half's tokens from one window and the second half's tokens from its successor (\Cref{fig:tokenization}). For synthesis, fixed token windows, each conditioned on a short speaker prompt (3s), are synthesized into waveform independently then merged with the same boundary overlap adjustment (\Cref{fig:synthesis}). We find these strategies minimize boundary artifacts while enabling continuous tokenization and decoding over time.

\textbf{Avoiding Implicit EOSes.} Despite having no end-of-sequence (EOS) tokens, our early models did not generatively extrapolate (e.g., a 4min model reaching 4.5min before degrading to noise/silence). In non-causal semantic tokenizers like USM-v2, we found the remaining length-in-window may be implicitly encoded in tokens, making tokens in final windows look ``different.'' As evidence, padding the last window to 30s using silence, tokenizing, then dropping those tokens led to silence, as ``future'' silence was now in the kept tokens. What worked was (1) to pad the last window to 30s using speech from the beginning of the example (depicted in \Cref{fig:tokenization}) so that final tokens were tokenized \text{as if} there was further speech, and (2) in the case of LibriLight, to still drop the last 10s of examples---as the padded beginnings were disproportionately "Chapter \textit{<number>}"!

\section{Improved Evaluations for Spoken LMs}
\label{sec:updated-nlg}
\label{sec:quality-over-time}

\textbf{Updated NLG Evaluations.} The shortcomings found in recent work (\Cref{sec:related-work-evals}) align with recent developments in natural language generation (NLG) evaluation, which have moved beyond intrinsic and/or surface word metrics like PPL, auto-BLEU, self-BLEU \citep{zhu2018-selfbleu}, especially for open-ended generation. One major shift has been the adoption of \textbf{embedding-based metrics}, where distances are computed between embeddings of generated versus reference text \citep{sai2022survey}. A more recent trend uses instruction-tuned LLMs to perform automated Likert-scaled evaluations \citep{li2024leveraging}, as applied by \citet{zhang2023speechgpt, zhang2024speechagents} to text-instructed speech generation and could be extended to speech as well.

However, to tackle the saturation of acoustic evaluations and to leverage text references, we in particular propose \textbf{automated side-by-sides}  (LLM-as-a-Judge; \citealp{zheng2023-llm-as-a-judge}) of total transcripts to scalably compare systems against the ground truth and each other. This has particular advantages for spoken LMs: (1) It mitigates the noise from ASR issues highlighted by \citet{hassid2023textually}, as the compared generations will be both afflicted (the \textit{ground truth should also be re-transcribed by ASR}, for fairness). (2) It works around the subtle issue of fixed-duration slices occurring mid-word,
degrading PPLs; instead, one can always transcribe prompt and continuation together, leaving the LLM to focus on the contrast. We implement both proposed evaluation types in Sections~\ref{sec:proposed-semantic}~and~\ref{sec:long-form-results}.

\textbf{LibriSpeech-Long.} To extend reference-based metrics to long-form speech, one needs long-form reference speech and transcripts. Although 3s prompts from LibriSpeech \citep{panayotov2015librispeech} have been the standard benchmark for spoken LMs since \citet{lakhotia2021generative}, their provided references have an average length of 10s, making them unsuitable beyond 7-10s continuations. Observing that the LibriSpeech dev and test sets are derived from full chapters of public-domain audiobooks which have been excluded from the standard LibriLight training set \citep{kahn2020libri} used by many spoken LMs, we reprocessed their source uncut audio files, similar to the convenience script\footnote{\href{https://github.com/facebookresearch/libri-light/blob/main/data_preparation/cut_by_vad.py}{libri-light/data\_preparation/cut\_by\_vad.py} on GitHub.} in LibriLight which agglomerates utterances up to a target length of 4 minutes (240s) along utterance boundaries. The results enable longer prompts and references for our side-by-side and similarity metrics. Statistics are shown in Table \ref{table:statistics}; 64\%--76\% of each split's examples are >3min long.

\begin{table}
\vskip -0.1in
\caption{Statistics of our proposed LibriSpeech-Long benchmark, which was generated with a maximum target duration of 4 minutes.}
\label{table:statistics}
\vskip 0.1in
\small
	\renewcommand{\tabcolsep}{1mm}
\centering
\resizebox{\linewidth}{!}{
\begin{tabular}{@{}lccccc@{}}
\toprule
\textbf{Subset}               & \# Hours  & \# Examples & Avg.\ Dur.\ (s) & \# Chapters  & \# Spkrs \\
\midrule
\addlinespace[0.5mm] 
dev-clean & 16.0  & 295 & 194.8 & 97 & 40 \\
dev-other  & 9.5 & 188 & 182.4 & 91 & 33             \\
test-clean   & 14.8  & 270 & 197.6 & 87 & 40             \\
\ \ \textit{>3.5min} & \textit{12.6} & \textit{193} & \textit{234.2} & \textit{82} & \textit{40} \\
test-other & 10.7  & 207 & 185.9 & 90 & 33             \\ 
\ \ \textit{>3.5min} & \textit{8.2} & \textit{126} & \textit{234.4} & \textit{77} & \textit{32} \\
\bottomrule
\end{tabular}
}
\vspace{-0.5cm}
\end{table}

\textbf{Generation Quality over Time.} While ASR can capture degenerate cases like repeated words (\Cref{fig:hero-transcript}), we found that it can fail on cases exacerbated by long-form generation, such as extended silences and voiced non-speech, suggesting the continued importance of audio-native evaluations. Furthermore, we find that generation failures generally increase as decoding progresses over time, which we must quantify to determine if our model has the desired property of generative length extrapolation (\Cref{sec:unbounded-speech-gen}). Towards this, we propose computing \textbf{semantic and acoustic metrics that are stratified over the decoding process}. This progression can be expressed in terms of semantic content (number of words into the ASR transcript), or acoustic content (time offset into the generated speech). We describe our implementations of both in \Cref{sec:proposed-longform}.

\begin{table*}
\vskip -0.1in
\centering
\caption{Results on short-form generation on LibriSpeech test-clean. Generations are 7s continuations of 3s prompts. \textbf{Bolded} are ours. Text-PT, FT denote pretraining (via LM init.) and finetuning with text. Win\%$_\text{GT}$ denotes the win rate of the model over the ground truth. $|V_{\text{audio}}|$ denotes speech token vocabulary size. For naturalness mean opinion score (N-MOS) we report 99\% confidence intervals.}
\vskip 0.1in
\small
\resizebox{\linewidth}{!}{
\begin{tabular}{@{}lcccccccccc@{}}
\toprule
\multirow{2}{*}{\textbf{Method}} & 
\multicolumn{5}{c}{\textbf{Text-Based (ASR)}} & 
\multicolumn{5}{c}{\textbf{Speech-Based}} \\ \cmidrule(lr){2-6} \cmidrule(lr){7-11}

& \textit{Text-PT} & \textit{Text-FT} & PPL$\downarrow$ & SBERT$\uparrow$ & Win\%$_\text{GT}\uparrow$ & \textit{$|V_{\text{audio}}|$} & SpkrSim$\uparrow$ & N-MOS$\uparrow$ & sWUGGY$\uparrow$ & sBLiMP$\uparrow$ \\ 
\midrule

{GSLM} (0.2B) & \ding{55} & \ding{55} & 6.28 & 0.17 & 1.4 & 100 & 0.36 & 2.23 ± 0.11 & 64.8 & 54.2 \\

{AudioLM (0.9B)} & \ding{55} & \ding{55} & -- & -- & -- & 1k & -- & -- & 71.5 & \textbf{64.7} \\

{TWIST-1.3B} & \ding{51} & \ding{55} & 7.25 & 0.18 & 3.6 & 500 & 0.41 & 3.09 ± 0.12 & 72.7 & 57.0  \\

{TWIST-7B} & \ding{51} & \ding{55} & 6.54 & 0.20 & \textbf{15.5} & 500 & 0.41 & 3.24 ± 0.13 & \textbf{73.9} & 59.0 \\

{VoxtLM (1.3B)} & \ding{51} & \ding{51} & -- & -- & -- & 200 & -- & -- & 65.6 & 57.1 \\

{Spirit LM Expressive (7B)} & \ding{51} & \ding{51} &  6.17 & 0.19 & 7.7 & 665 & 0.45 & 3.00 ± 0.08 & 65.0 & 54.2  \\

\midrule

{\textbf{SpeechSSM-2B}} & \ding{51} & \ding{55} & 5.76 & \textbf{0.23} & 7.9 & 32k & \textbf{0.79} & 3.87 ± 0.07 & 55.8 & 60.9 \\
\quad\textit{without LM init.} & \ding{55} & \ding{55} & 6.15 & \textbf{0.23} & 7.7 & 32k & \textbf{0.79} & 3.80 ± 0.07 & -- & -- \\
\quad\textit{with Transformer instead} & \ding{51} & \ding{55} &  6.16 & 0.22 & 8.4 & 32k & \textbf{0.79} & 3.74 ± 0.08 & -- & -- \\
\quad\textit{with 30s segs.\ instead} & \ding{51} & \ding{55} & 5.73 & 0.22 & 10.5 & 32k & \textbf{0.79} & 3.84 ± 0.08 & 57.3 & 61.1 \\
\quad\textit{with 16min segs.\ instead} & \ding{51} & \ding{55} & 5.84 & 0.20 & 4.0 & 32k & \textbf{0.79} & {3.86 ± 0.07} & 54.3 & 60.4 \\
{\textbf{SpeechSSM-9B}} & \ding{51} & \ding{55} & \textbf{5.60} & \textbf{0.23} & 13.5 & 32k & \textbf{0.79} & \textbf{3.94 ± 0.07} & -- & --\\

\midrule

\textit{Ground Truth} & -- & -- & \textit{5.63} & \textit{1.00} & \textit{50.0} & -- & \textit{0.84} & \textit{4.02 ± 0.07} & -- & -- \\

\bottomrule
\end{tabular}
}
\label{table:shortform_eval}
\label{table:swuggy_sblimp}
\end{table*}

\section{Experimental Setup}
\label{sec:experimental-setup}

\textbf{Training and Generation.} Following \citet{borsos2023audiolm}, \citet{nachmani2023spoken}, and others, we train on the unlab-60k split from LibriLight \citep{kahn2020libri}. Unlike prior work, we study the effect of sequence length on long-form generation, segmenting the audiobooks into training sequences of up to a target duration. The default for  \textbf{SpeechSSM} is 4min (240s) during training, though we compare with target durations of 30s and 16min (960s) as well (``\textit{with 30s/16min segments}''). We train \textbf{-2B} and \textbf{-9B} variants, corresponding to RecurrentGemma. Each model is trained with 16 TPUs (v5p) and data parallelism for 100k steps with 768k tokens per batch, and a checkpoint is chosen via transcript perplexity on LibriSpeech-Long dev-clean; more details in \Cref{sec:further-training}. We sample semantic (USM-v2) tokens with temperature 1. Then the SoundStorm model (speaker-prompted with the first 3 seconds of the prompt) and windowing (30s with 4s overlaps) give acoustic (SoundStream) tokens, which a SoundStream codec turns into waveform; model details for these are in \Cref{sec:further-model-details}.

\textbf{Baselines.} We compare SpeechSSM with decoder-only speech LMs. We use \textbf{GSLM} \cite{lakhotia2021generative}'s best model (HuBERT-L6 tokens with vocab 200), trained on the clean 6k hours of LibriLight. For \textbf{TWIST} \cite{hassid2023textually}, we use both the OPT-1.3B and LLaMA-7B versions, trained on 150k speech hours. For the 7B \textbf{Spirit LM} \citep{nguyen2024spirit}, we use the \textbf{Expressive} version model which adds expressiveness via pitch and style tokens in addition to HuBERT tokens. We also cite numbers from \textbf{AudioLM} \citep{borsos2023audiolm} and \textbf{VoxtLM} \citep{maiti2023voxtlm}, which are both 2B models trained on unlab-60k. VoxtLM and Spirit LM see text data during training. Due to variations in token, initialization, and training data choices, we also define \textbf{SpeechTransformer} (``\textit{with Transformer}''), a spoken LM initialized with Gemma-2B \citep{team2024gemma} but otherwise matched with SpeechSSM-2B.\footnote{Note that Gemma saw 50\% more text than RecurrentGemma.}

\section{Short-Form Continuation Experiments}
\label{sec:gen-metrics}

Before considering long-form, we compare SpeechSSM to its Transformer-based counterparts as in past work. This takes 3s prefixes from LibriSpeech's test-clean set and generates 7s continuations, which one then transcribes with an ASR model (our details in \Cref{sec:further-evaluation}). We used test examples with ground-truth continuations $\ge$5s.

\subsection{Existing Metrics and Results}
\label{sec:existing-short}

\textbf{Transcript Perplexity (PPL):} As in prior work, we compute the log-perplexity of the transcript of the generated continuation under Gemma-2B, as an initial proxy for content fluency.

\textbf{Speaker Similarity (SpkrSim):} To analyze voice preservation, we speaker-embed both the prompt and its generated continuation and compute their cosine similarity. We use a speaker classifier used in AudioLM \cite{borsos2023audiolm} as the speaker embedder.

\textbf{Naturalness Mean Opinion Score (N-MOS):} We evaluate how natural the speech sounds, ignoring the grammar and content of the utterance; this focuses attention on issues not visible on transcripts, ranging from synthesis issues and unintelligible speech through to coherent but unnatural prosody; rater details in \Cref{sec:mos-instruction}.

\textbf{sWUGGY and sBLiMP:} These probe if the spoken language model can implicitly perform lexical and syntactic contrasts \citep{nguyen2020zero}.\footnote{sWUGGY: audio pairs of a real word and a fake but similar-sounding word. sBLIMP: audio pairs of a syntactically correct sentence and an incorrect one.} One reports the \% of time the model's log-likelihood ranks the semantic tokens of a correct utterance versus its incorrect counterpart.

Our models' continuations (Table \ref{table:shortform_eval}) are most speaker-similar to the prompt, which we attribute to our speaker-promptable acoustic stage; in contrast, GSLM and TWIST can only propagate speaker identity via their semantic tokens (plus coarse style and pitch tokens in Spirit LM Expressive). Our N-MOS scores suggest high semantic-to-audio synthesis quality, which we attribute to USM-v2's large vocabulary (32k) and the staged approach via existing codec (SoundStream). SpeechSSM's naturalness is on par or better than a comparable Transformer and close to real speech; see supplementary.
Meanwhile, our sWUGGY scores are much worse, while our sBLiMP scores are neutral to above-average. These do not positively correlate with text-based scores or subjective quality; instead, they match \citet{lakhotia2021generative}'s observations and \citet{borsos2023audiolm}'s Figure 2, where sWUGGY scores hit relatively sharp maxima at vocabularies of a few hundred tokens. We argue increased noise is expected with larger vocabularies, as then tokens of an audio utterance represents less of the probability mass of all possible renditions of its text. Hence, separately from \citet{defossez2024moshi}, we move towards transcript-based evaluations, as these \textit{marginalize} over spoken renditions to get less noisy evaluations.
Finally, even the 2B SpeechSSMs had the lowest transcript perplexities, which is surprising given the other models are larger (7B) and/or have jointly trained with text, and one would not expect SSMs to confer a semantic edge in the $\le$10s speech horizon. We note that \citet{lakhotia2021generative} already caveated---and \citet{hassid2023textually} actively discouraged---the use of ASR PPL given its sensitivity to e.g.\ audio sampling temperature; such metrics may simply indicate model repetitiveness at default settings.

\begin{table*}
\vskip -0.1in
\caption{Evaluations on LibriSpeech-Long test-clean. $\boxplus$ denotes model extension via windowed generation. Generations are 4m completions of the 10s prompt. Win\%$_\text{SSM-2B}$ and Win\%$_\text{GT}$ are model wins versus SpeechSSM-2B and ground truths (>3.5min) respectively. For naturalness MOS over time (N-MOS-$T$), the same 5s time span is sampled over all models from each minute of completion.}
\label{table:longform_eval}
\label{table:mos}
\vskip 0.1in
\small
\centering
\resizebox{\linewidth}{!}{
\begin{tabular}{@{}lccccccccc@{}}
\toprule
\multirow{3}{*}{\textbf{Method}} &
\multicolumn{4}{c}{\textbf{Text-Based (ASR)}} & 
\multicolumn{5}{c}{\textbf{Speech-Based}}
\\ \cmidrule(lr){2-5} \cmidrule(lr){6-10}

& \multirow{2}{*}{PPL$\downarrow$} & \multirow{2}{*}{Gecko$\uparrow$}  &  \multirow{2}{*}{Win\%$_\text{SSM-2B}\!\uparrow$}  &  \multirow{2}{*}{Win\%$_\text{GT}\!\uparrow$}  & \multirow{2}{*}{SpkrSim$\uparrow$} & \multicolumn{4}{c}{N-MOS-$T$\,$\uparrow$ (99\% CI)} \\ \cmidrule(lr){7-10}

&  &  &   &  & & $<$1min  & 1-2min  & 2-3min  & $\ge$3min \\ \midrule

{GSLM (0.2B)}$^\boxplus$ &  4.74 & 0.67 & 17.6 & 0.0 & 0.33 & 2.00 ± 0.22 & 2.01 ± 0.23 & 1.93 ± 0.21 & 2.09 ± 0.22 \\

{TWIST-1.3B}$^\boxplus$ & 5.70 & 0.60 & 16.2 & 0.0 & 0.38 & 1.96 ± 0.22 & 1.59 ± 0.13 & 1.40 ± 0.09 & 1.36 ± 0.09\\

{TWIST-7B}$^\boxplus$ & 4.93 & 0.65 & 24.0 & 0.0 & 0.37 & 2.43 ± 0.27 & 2.15 ± 0.22 & 1.83 ± 0.20 & 1.84 ± 0.18 \\

{Spirit LM Expressive (7B)} &  5.71 & 0.63 & 24.2 & 0.0 & 0.41 & 2.77 ± 0.12 & 2.52 ± 0.12 & 2.56 ± 0.11 & 2.60 ± 0.10  \\

\midrule

{\textbf{SpeechSSM-2B}} & 3.75 & 0.70 & \textit{50.0} & 0.0 & \textbf{0.85} & \textbf{4.12} ± \textbf{0.08} & \textbf{4.13} ± \textbf{0.06} & \textbf{4.13} ± \textbf{0.07} & \textbf{4.16} ± \textbf{0.08}   \\
\quad\textit{without LM init.} & 4.57 & \textbf{0.71} & 42.4 & 0.0 & \textbf{0.85} & 3.76 ± 0.09 & 3.76 ± 0.10 & 3.71 ± 0.10 & 3.80 ± 0.08 \\
\quad\textit{with Transformer instead} &  4.71 & \textbf{0.71} & 33.9 & 0.0 & 
\textbf{0.85} & 4.03 ± 0.09 & 3.88 ± 0.08 & 3.98 ± 0.09 & 3.89 ± 0.12 \\
\quad\textit{with 16min segments instead} & 3.83 & 0.70 & 46.1 & 0.0 & \textbf{0.85} & 3.83 ± 0.08 & 3.89 ± 0.08 & 3.87 ± 0.09 & 3.88 ± 0.09 \\

{\textbf{SpeechSSM-9B}} & \textbf{3.57} & \textbf{0.71} &  \textbf{75.0} & 0.0 & \textbf{0.85} & 3.86 ± 0.08 & 3.90 ± 0.08 & 3.88 ± 0.07 & 3.95 ± 0.07 \\

\midrule

\textit{Ground Truth} & \textit{3.61} & \textit{1.00} & \textit{100.0} & \textit{50.0} & \textit{0.92} & \textit{4.12} ± \textit{0.08} & \textit{4.11} ± \textit{0.08} & \textit{4.10} ± \textit{0.08} & \textit{4.09} ± \textit{0.08}   \\

\bottomrule
\end{tabular}
}

\end{table*}

\subsection{Proposed Metrics and Results}
\label{sec:proposed-semantic}

The above points---noise from contrastive audio probes, saturation of N-MOS and SpkrSim, suspicious results from transcript perplexity---\textbf{all motivate our proposed shift to newer, reference-based NLG metrics} (\Cref{sec:updated-nlg}):

\textbf{Semantic Similarity (SBERT).} We measure the distance between the semantic embedding of the transcriptions of the generated speech and the reference, using Sentence-BERT MiniLM-L6-v2 \cite{reimers2019sentence} as the semantic embedder. This expresses contextual alignment between the generated text to the ground truth, focusing on semantic meaning over surface-form patterns.  

\textbf{Side-by-Side Win Rates (Win\%$_\text{vs.\,model}$).} We ask the model to analyze \text{then} rate \citep{chiang2023-analyze-rate}, forking the format of Arena-Hard-Auto's LLM-Judge System Instruction \citep{li2024-arena-hard}. Given the book domain and the relatively unconstrained nature of speech continuation, we base our criteria on questionnaires from story generation evaluation \citep{xia2023-storytelling} around \textit{fluency}, \textit{coherence}, \textit{logicality}, and \textit{interestingness}; see \Cref{sec:llm-judge-prompt} for the template. We evaluate each prompt twice with order of presentation flipped. Gemini 2.0 Flash \cite{team2024gemini} re-transcribes both model and true audio (without windowing and jointly with prefix) and performs judgments. 

Benefits are evident in side-by-sides versus a transcript of the ground truth, with results (Table \ref{table:shortform_eval}) now matching qualitative experience and expectations (larger models performing better, with Spirit LM Expressive underperforming TWIST perhaps due to capacity spent on pitch/style). However, even SpeechSSM-9B and TWIST-7B win <20\% versus transcribed ground truth, suggesting that (automated) \textbf{side-by-side comparison on transcripts is more discriminative than a holistic human side-by-side audio task} in selecting the synthetic sample, where humans performed at random in \citet{borsos2023audiolm}, as it focuses on the content of the speech rather than superficial naturalness. Though our model is not the most fluent in this regime, SBERT foreshadows benefits in faithfulness (our models' continuations are semantically closest to the true ones).

\section{Long-Form Generation Experiments}
\label{sec:long-form-exps}

We conduct the first evaluation of long-form speech generation, taking extended prompts of 10s from our proposed LibriSpeech-Long (test-clean) and having each model continue them through to 4 and 16 minutes. As other off-the-shelf models trained on sequences well below 4min and were not designed to generate beyond their training length (e.g., use of position encodings), we found them unable to generate beyond a minute without being stuck in noise or silence. To give functional baselines, we propose applying \textit{slide-and-prompt} generation \cite{borsos2023soundstorm} to the semantic LM itself; that is, we generate to each model's maximum completion length (\Cref{fig:prior_work}) first, and then repeat generation using the last 3s of the previous window as context. Example generations are in \Cref{sec:further-generations}.

\subsection{Semantic and Acoustic Results}
\label{sec:long-form-results}

We again measure existing and our proposed metrics, with two key changes: first, \textit{we replace Sentence-BERT with Gecko} \citep{lee2024gecko} as embedder for semantic similarity as Sentence-BERT's 512-token context cannot handle the transcripts of 4min+ generations; more long-form evaluation details in \Cref{sec:further-evaluation}. Second, extrapolation failures cause e.g., GSLM and TWIST to generate far less than our models; for fairness \textit{in side-by-sides, we truncate to the shorter transcript's length}.\footnote{Note this means the \textit{relative order} of models in a single Win\%$_\text{vs.\,model}$ column should not be read into too much, as challenger models induce truncation to very different lengths.}

Our results are in Table~\ref{table:longform_eval}. The SpkrSims of SpeechTransformer, SpeechSSM, and ground truth increased from 0.79 in short-form to 0.85, likely from increased confidence from longer prompt and continuations; all other decreased. This shows the advantage of our more speaker-invariant USM-v2 tokens and a speaker-prompted audio stage; identity is modulated by the semantic-to-acoustic model, instead of consuming capacity, and being imperfectly carried by, the semantic LM and windowing. SpeechSSM has the best ASR PPL and wins a majority of time vs.\ all models; our 2B Transformer variant is close at 30sec, but not at 16min (Table~\ref{table:16min_eval}) ---recall that 16min is \textit{24,000} USM-v2 tokens!

However, models achieve \textit{zero wins over the ground truth}. As generation length increases, faults in fluency, coherence, logicality and interestingness become increasingly apparent, showing that \textbf{side-by-side comparison versus LibriSpeech-Long ground-truths is a new and difficult benchmark} for (long-form) spoken language generation.

\subsection{Proposed Extrapolation Metrics and Results}
\label{sec:proposed-longform}

\begin{table}[t]
\vskip -0.1in
\caption{16min completions of 10s prefixes of our LibriSpeech-Long test-clean. As there are no 16min ground truths, we only take reference-free metrics. $\boxplus$ denotes model extension via slide-and-prompt. Win\%$_\text{SSM-2B}$ denotes win rate over SpeechSSM-2B.}
\label{table:16min_eval}
\vskip 0.1in
\small
\centering
\resizebox{0.8\linewidth}{!}{
\begin{tabular}{@{}lcc@{}}
\toprule
\multirow{1}{*}{\textbf{Method}} & \multirow{1}{*}{PPL\,$\downarrow$} & \multirow{1}{*}{Win\%$_\text{SSM-2B}\!\uparrow$} \\ \midrule

{GSLM (0.2B)}$^\boxplus$ & 4.48 & 18.4 \\

{TWIST-1.3B}$^\boxplus$ & 4.76 & 17.5 \\

{TWIST-7B}$^\boxplus$ & 4.45 & 32.8 \\

{Spirit LM Expressive (7B)}$^\boxplus$ & 5.66 & 37.7 \\

\midrule

{\textbf{SpeechSSM-2B}} & 3.59 & \textit{50.0} \\
\quad\textit{with Transformer instead} & 5.91 & 24.9 \\
\quad\textit{with 16min segments instead} & 3.55 & 48.4 \\
{\textbf{SpeechSSM-9B}} & \textbf{3.39} & \textbf{68.3} \\

\bottomrule
\end{tabular}
}
\vspace{-0.3cm}
\end{table}

\textbf{Naturalness Mean Opinion Score over Time (N-MOS-$T$).} To balance cost and informativeness, for each example we select a 5 sec.\ span from each minute; specifically [$t_\text{prompt}$, 60), [60, 120), [120, 180), and [180, $t_\text{max}$) where $t_\text{prompt}$ is prompt duration and $t_\text{max}$ is ground truth duration, and extract each span's audio from every model's generated continuation.

\textbf{Semantic Coherence over Length (SC-$L$).} To evaluate semantic faithfulness to the original prompt over time while mitigating the effect of speech rate, we take each continuation's transcripts and divide them into spans of 100 words (as determined by whitespace). Each 100-word segment represents $\sim$30 seconds of speech, with the advantage of normalizing out differences in generated speech rate as well as degenerate silences. Our SC scores are cosine similarities between the embedding of the original prompt $\mathbf{e}_\text{prompt}$ with that of each 100-word segment $\mathbf{e}_{100\ell : 100(\ell + 1)}$.

\begin{figure}[t]
        \includegraphics[width=\linewidth]{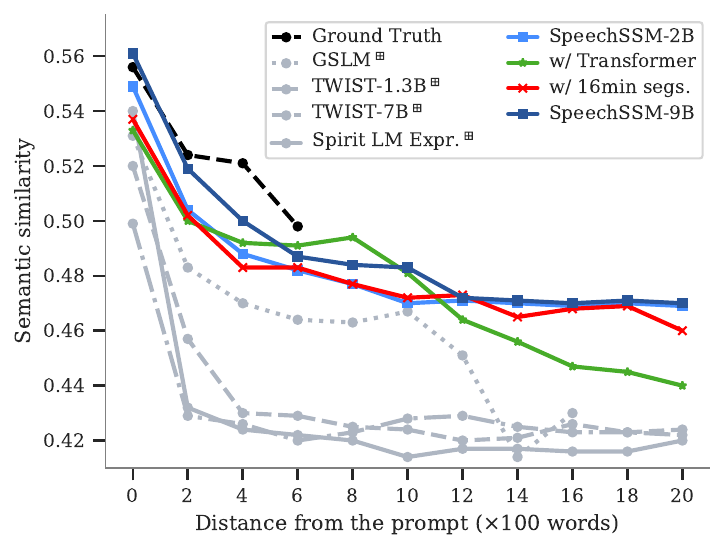}
        \vspace{-0.5cm}
        \caption{Semantic similarity between a 10s prompt and the 100-word segment starting at word 100$L$ in the 16min completion, as measured by cosine similarity of Gecko embeddings (SC-$L$).}
        \label{fig:SC-L-16min}
\end{figure}

These metrics, motivated in \Cref{sec:quality-over-time} and shown in Tables~\ref{table:longform_eval}~and~\ref{table:16min_eval}, show that \textbf{windowed model extension is a poor length extrapolator}. GSLM and TWIST are already low in our proposed N-MOS-$T$ by minute one, having trained on even shorter sequences. Spirit LM, which has seen sequences up to 200s (though rarely, due to text interleaves), degrades acoustically over time, though slower. 

Our proposed SC-$L$ is plotted for 16 minutes in \Cref{fig:SC-L-16min} (table in \Cref{sec:further-results}).
As generation progresses, we see a decline in SC-$L$ scores, aligning with the natural flow of speech starting on a topic and evolving over time. However, existing models except GSLM sharply drop in semantic coherence (SC) scores around 200 words ($\sim$1min; around their training lengths). GSLM fares better as its failures are untranscribed noise which do not show in SC-$L$ (but do in N-MOS-$T$), but still worse than our models which are closest to the 3-4min ground truths' performance.
At 16min (\Cref{table:16min_eval}, \Cref{fig:SC-L-16min}), we do not see obvious degradation from SpeechSSM-2B generating beyond its training length.
In contrast, SpeechTransformer-2B had comparable PPL, win-rate, and SC scores up to its training length, but these metrics quickly degrade at 16min. This \textbf{suggests an edge for SSMs in long-form speech generation}, complementing past work \citep{gu2021efficiently}. In all, SpeechSSM's design and architecture lead to generative length extrapolation.

\subsection{Qualitative Discussion}

We invite the reader to listen to the samples on our website. For convenience, Figure \ref{fig:hero-transcript} shows that even SpeechSSM-2B generates intelligible and more coherent speech over the 4min duration, with ongoing references to the "Philip" in the prompt, along with new recurring entities like "Prince Maria", "Prince Albert", and "Horace Barrows" also consistently appear, maintaining contextual relevance. We provide further transcripts for ours as well as other models in \Cref{sec:further-generations}, showing how their decline in SC-$L$ is qualitatively visible as e.g., degeneration into repetitive outputs.

Given our use of windows during synthesis (\Cref{sec:unbounded-speech-gen}), one may wonder if audio quality differs at transitions (every 25 + 23$n$ seconds; the multiple comes from 30s, minus 3s for the prompt and 2s truncation per side). We ran human evaluation comparing 5-second windows centered at chunk boundaries vs.\ chunk midpoints, and found no clear effect on MOSes or six categories of rater-flaggable issues (\Cref{sec:boundary_artifacts}), which matches our informal listening.\footnote{That said, targeted listening does reveal e.g., subtle loudness shifts at times; still a massive reduction versus synthesizing without overlaps or without a fixed prompt.}

\subsection{Inference Efficiency}

\begin{figure}[t]
        \includegraphics[width=0.9\linewidth]{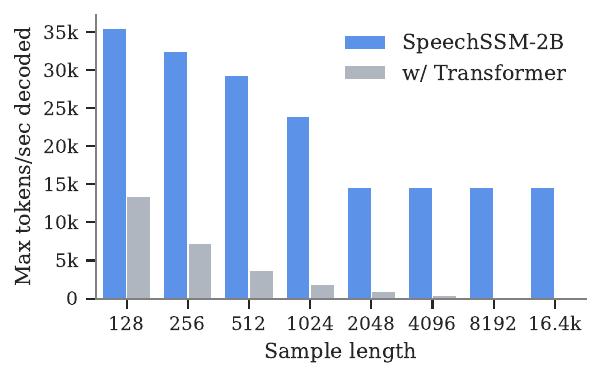}
        \vspace{-0.4cm}
        \captionof{figure}{Max throughput under batch decoding per model and sampling length on one TPU v5e (unconditional generation).}
        \label{fig:throughput}
\end{figure}

\begin{figure}[t]
        \includegraphics[width=0.9\linewidth]{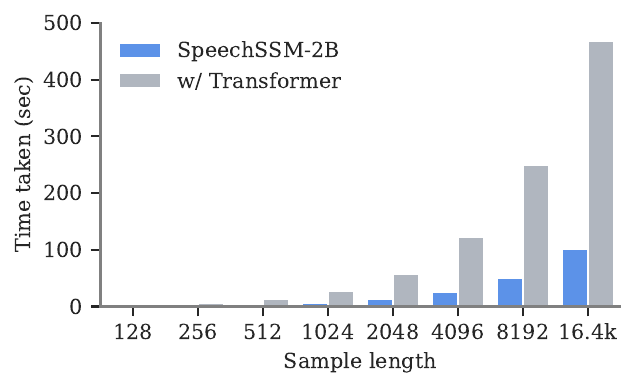}
        \vspace{-0.4cm}
        \captionof{figure}{Time taken to decode a single sample (batch size 1) to a target length on one TPU v5e (unconditional generation).}
    \label{fig:sampling-time}
\end{figure}

\textbf{Throughput} measures the maximum number of tokens that can be successfully decoded per second given fixed memory, e.g.,\ by increasing the batch size. In \Cref{fig:throughput}, SpeechSSM attains higher throughput due to its recurrent layers that maintain a constant-size state; furthermore, its self-attention has a maximum context of size 2048, giving a lower bound. In all, this confirms SpeechSSM's capped memory use and thus capacity for unbounded generation, with >120x the throughput of SpeechTransformer once decoding 16.4k-token sequences in batch (also depicted in \Cref{sec:throughput_speedup}).
\textbf{Decoding time} measures the speed at which a sample can be decoded to a desired length \Cref{fig:sampling-time}. We see that SpeechSSM also increases speed per step versus SpeechTransformer; an additional factor in SpeechSSM's improved throughput. On the TPU v5e, the 2B SpeechSSM decodes 16.4k tokens (roughly 10.9 minutes) in just over 100 seconds, a real-time factor well under 0.2x.

\subsection{Extemporaneous Speech Generation}

With few exceptions (dGSLM, Spirit LM Expressive), most spoken LMs are trained on read speech such as LibriLight. However, long-form multimedia and assistant applications likely require modeling of spontaneous speech, which has its own long-term discourse structures (e.g., podcasts; \citealp{nishimura2024halle}). Hence, we also developed \textbf{SpeechSSM-X}, a 2B SpeechSSM model trained on a 216k-hour corpus of eXtemporaneous monologues (see \Cref{sec:speechssm-x-details} for details). We find that it is able to generate natural monologue speech in a more informal, extemporaneous style, while showing similar coherence at the multi-sentence level; see our website for examples.

\section{Conclusion}

We considered the task of generative modeling of long-form speech. For modeling, this led us to SpeechSSM, the first spoken state-space LM, allowing generation than can go indefinitely without running out of memory. For evaluation we created the LibriSpeech-Long benchmark and proposed new evaluations for long-form speech continuation.
We hope our work will simplify and enable new audio generation applications involving long-form media, such as audiobooks, podcasts, voice agent sessions, and video-related content.

\section*{Acknowledgements}
We are grateful to Zalán Borsos for their technical feedback; to Tongzhou Chen, Roshan Sharma, and members of our Speech and Griffin teams for infrastructure support; and to Soroosh Mariooryad for general assistance. We also thank the anonymous reviewers and area chair for their support and helpful feedback.

\section*{Impact Statement}




Progress on coherent, efficient, and unbounded audio-native speech language models will improve the availability and accessibility of audio-based human-computer interfaces, particularly towards the generation of speech with paralinguistic nuances often inexpressible in a standard text-to-speech prompt, as well as enabling models in primarily oral languages (textless NLP). It should also stimulate the use of synthetic speech in long-form creative multimedia applications, and could conceivably generalize to non-speech audio such as music.

We recognize that increased efficiencies in speech language modeling may increase the proliferation of audio deepfakes or low-quality synthetic media. However, for these uses we do not believe our work exacerbates what is already achievable (at higher content quality) by cascading text large language models (LLMs) with modern voice-promptable text-to-speech systems. We hope our work increases public awareness that direct methods in deep learning now enable large-scale generation of expressive speech, analogous to the situation in text media due to LLMs.

As in prior work for generative speech language modeling, speaker prompts used in our read-speech demos come from public domain LibriVox audiobooks and are only used to generate spoken continuations in the same setting, albeit for longer durations. For our spontaneous-speech demos, the voices have been explicitly licensed for speech synthesis. We do not currently release model weights.

\bibliography{custom}
\bibliographystyle{icml2025}

\newpage
\appendix

\section{Additional Results}
\label{sec:further-results}

\subsection{SC-$L$ (16 min.)}
\label{sec:further-sc-16min}

\begin{table}[!ht]
\caption{Semantic coherence scores are over the indicated spans of words for the transcripts of 16-minute completions. $\boxplus$ denotes model extension via windowed generation. GSLM's blank is due to not generating that many words in the time period. Ground Truth's blanks are due to its audio being $\le$4min.}
\label{table:16min_sc}
\vskip 0.1in
\small
\centering
	\renewcommand{\tabcolsep}{1mm}
\resizebox{\linewidth}{!}{
\begin{tabular}{@{}lcccc@{}}
\toprule

\multirow{2}{*}{\textbf{Method}} & \multicolumn{4}{c}{SC-$L$\,$\uparrow$} \\ \cmidrule(lr){2-5}

 & 0-100  & 600-700  & 1200-1300  & 1800-1900 \\ \hline

{GSLM (0.2B)}$^\boxplus$ & 0.531 & 0.464 & 0.451 & -- \\

{TWIST-1.3B}$^\boxplus$ & 0.499 & 0.420 & 0.429 & 0.423\\

{TWIST-7B}$^\boxplus$ & 0.520 & 0.429 & 0.420 & 0.423\\

{Spirit LM Expressive (7B)}$^\boxplus$ & 0.540 & 0.422 & 0.417 & 0.416 \\

\midrule

{\textbf{SpeechSSM-2B}} & 0.549 & 0.482 & 0.471 &  0.470 \\
\quad\textit{with Transformer instead} & 0.533 & \textbf{0.491} & 0.464 & 0.445  \\
\quad\textit{with 16min segments instead} & 0.537 & 0.483 & \textbf{0.473} & 0.469  \\
{\textbf{SpeechSSM-9B}} & \textbf{0.561} & 0.487 & 0.472 & \textbf{0.471} \\

\midrule

{\textit{Ground Truth}} & \textit{0.556} & \textit{0.498} & -- & -- \\

\bottomrule
\end{tabular}
}
\end{table}

\subsection{Human Ratings around Synthesis Boundaries}
\label{sec:boundary_artifacts}

From SpeechSSM-2B's long-form continuations to 4 minutes, we take our subset of 50 prompts (\Cref{sec:further-evaluation}). For each, we take five-second windows centered at synthesis chunk boundaries versus synthesis chunk centers, giving 10 windows of each type. Each gets 2 ratings, to give 1,000 ratings per condition (\Cref{table:mos_concat_eval}). The mean opinion scores (MOSes) are nearly identical (4.05±0.07 vs.\ 4.07±0.07) and there is no clear trend in synthesis issues, suggesting concatenation boundaries do not lead to net loss in naturalness to an attentive but non-specialist listener.

\begin{table}[ht!]
\vskip -0.1in
\caption{Human evaluation of generation quality of SpeechSSM-2B at and away from chunk boundaries. We report Mean Opinion Scores (MOS) with 99\% confidence intervals along with rater-flagged error categories.}
\label{table:mos_concat_eval}
\vskip 0.1in
\small
	\renewcommand{\tabcolsep}{1mm}
\centering
\resizebox{\linewidth}{!}{
\begin{tabular}{@{}lccccccc@{}}
\toprule
\textbf{Location} & \textbf{MOS} & Artifacts & Pronunciation & Speed & Prosody & Sentiment & Other \\
\midrule
\addlinespace[0.5mm]
Chunk boundary & 4.05 ± 0.07 & 115 & 16 & 41 & 72 & 38 & 26 \\
Chunk center  & 4.07 ± 0.07 & 122 & 16 & 53 & 78 & 32 & 20 \\
\bottomrule
\end{tabular}
}
\end{table}

\subsection{Relative Throughputs}
\label{sec:throughput_speedup}

\begin{figure}[ht!]
    \centering
    \includegraphics[width=0.6\linewidth]{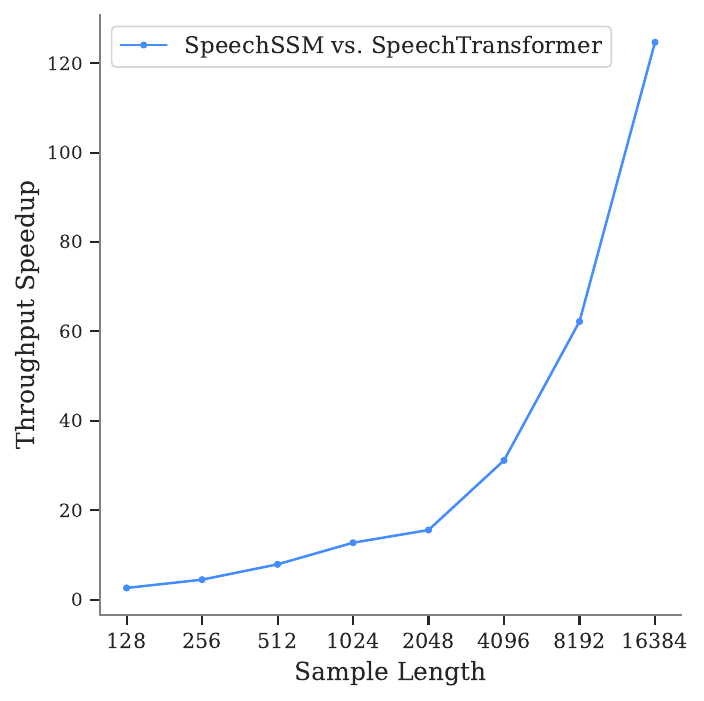}
    \vspace{-0.5cm}
    \caption{Ratio of SpeechSSM to SpeechTransformer's throughput on a single TPU v5e performing unconditional generation at different sampling lengths, based on \Cref{fig:throughput}.}
    \label{fig:throughput_speedup}
    \vspace{-0.4cm}
\end{figure}

In \Cref{fig:throughput_speedup} we compare SpeechSSM-2B's throughput versus SpeechTransformer.

\section{Additional Implementation Details}

\subsection{Model Training and Selection}
\label{sec:further-training}

SpeechSSM-2B w/ 30s, 4min, and 16min segments get 750, 5760, and 24k tokens per segment respectively, under USM-v2's frame rate of 25Hz. For the 4min's length extrapolation (\Cref{sec:unbounded-speech-gen}), we drop the last 10s of each segment. Each version is trained on groups of 512, 128, and 32 sequences respectively, amounting to 768k tokens per batch. We train with the Adam optimizer with a learning rate of 5e-4 and weight decay 0.6 for 100k steps (with a warmup of 1k steps and a cosine decay schedule to 1/20th the learning rate). We select the best model on a fixed random subset of LibriSpeech-Long dev-clean, by generating continuations with each model to its target length over 5 checkpoints, then choosing the one which minimizes transcript perplexity (\Cref{sec:existing-short}).

\subsection{SoundStorm and SoundStream}
\label{sec:further-model-details}

Our SoundStorm follows the original hyperparameters for a voice-promptable model as described in \citet{borsos2023soundstorm} (30s windows), except we double the number of layers to 24 to give 600M parameters, and our model was pretrained on a mixture of Common Voice v11.0 \citep{ardila2020common}, Multilingual LibriSpeech \citep{pratap2020mls}, and VoxPopuli \citep{wang2021voxpopuli}. This is comparable to other works which do not restrict their synthesizer data to LibriLight: TWIST \citep{hassid2023textually} does not explicitly indicate its vocoder data, but for HuBERT and the speech LM itself the English splits of these three datasets were used in training among many others; Spirit LM \citep{nguyen2024spirit} used Expresso \citep{nguyen2023expresso} to train its vocoder.

\subsection{Model Evaluation}
\label{sec:further-evaluation}

For text-based evaluations, we transcribe the generated speech to enable the application of natural language generation (NLG) metrics \citep{lakhotia2021generative}. Unless stated otherwise, we use wav2vec2-base-960h \cite{baevski2020wav2vec} for ASR, applied on 180s windows. To reduce cost and to mitigate length/duration as an indicator for ground truth, for more intensive evaluations (N-MOS, side-by-sides) we randomly selected 200 test-clean utterances with ground-truth continuations $\ge 7$s.

\textbf{For long-form evaluations:} Gecko is a text embedding model trained on a variety of tasks including document retrieval, semantic similarity, and classification using long passages, and is more suited for extracting semantic embedding of long texts in open-domain generation. The choice of task is given via prompting; we use `search results' as the prompt which was used to train on clustering tasks. For win-rates, to mitigate length bias we only consider the 193 examples $\geq$3.5min (71\% of test-clean). For MOS computations, we randomly selected 50 from these.

\subsection{SpeechSSM-X}
\label{sec:speechssm-x-details}

We use speaker-based detection to categorize audio files into long-form monologue content, with turns marked by voice activity detection. We take contiguous turns of up to 16min in length as training sequences.

\section{Additional MOS Evaluation Details}
\label{sec:mos-instruction}

For short-form N-MOS, we took our subset of 200 prompts (\Cref{sec:further-evaluation}) and collected 6 ratings for each model continuation. For N-MOS-$T$, we took our subset of 50 prompts and collected 24 ratings for each model continuation and slice. Hence, each score represents 1200 ratings unless stated otherwise.

The following prompt was used in all cases:\\

\begin{text-sample}
(placeholder since first line is not rendered for some reason)

This task requires you to listen to audio clips using headphones in a quiet environment.

In this task, you will be given audio clips. For each clip, please listen to the speech very carefully and then select a rating on a scale of 1 (very unnatural) to 5 (very natural) using 0.5 point increments. The rating should be based on how natural or unnatural the speech sounded. Please do not judge the grammar or the content of the sentence. Instead, just focus on how natural the speech sounds.

Possible Ratings
1: Bad
1.5
2: Poor
2.5
3: Fair
3.5
4: Good
4.5 
5: Excellent 

\end{text-sample}

\onecolumn

\section{LLM-as-a-Judge Example}
\label{sec:llm-judge-prompt}

This is our model prompt, with the example of comparing a transcript of the ground truth versus a transcript of GSLM's \citep{lakhotia2021generative} generation. This example was chosen to highlight the rater model's acknowledgement of the prompt request to not penalize incomplete sentences:

\begin{text-sample}
(placeholder since first line is not rendered for some reason)

# Instructions

Please act as an impartial judge and evaluate the quality of two texts which occur in the context of a book. These texts are transcribed from audio recordings that were truncated to a fixed duration. Your job is to consider the following criteria to evaluate which text is better:
- Fluency: How grammatically correct is the text?
- Coherence: How well do the sentences of the text fit together?
- Logicality: How much does the text obey common sense?
- Interestingness: How enjoyable is the text to read?

First, read text A and consider its fluency, coherence, logicality, and interestingness. Do not penalize the text for ending mid-sentence or mid-paragraph.

Then, read text B and consider its fluency, coherence, logicality, and interestingness. Do not penalize the text for ending mid-sentence or mid-paragraph.

Afterwards, compare the fluency, coherence, logicality, and interestingness of the two texts. Do not penalize either text for ending mid-sentence or mid-paragraph.

Finally, after providing your explanations, you must output only one of the following choices as your final verdict with a label:
1. Text A is significantly better: [[A>>B]]
2. Text A is slightly better: [[A>B]]
3. Tie, relatively the same: [[A=B]]
4. Text B is slightly better: [[B>A]]
5. Text B is significantly better: [[B>>A]]

Example output: "My final verdict is tie: [[A=B]]".

# Comparison task

## ---------- Text A ----------

Pearl accordingly ran to the bow window at the further end of the hall and looked along the vista of a garden walk carpeted with closely shaven grass and bordered with some

## ---------- Text B ----------

Pearl accordingly ran to the bow window at the further end of the table. "Palmer to Austin and two boys," said the next fore keeper.

## ---------- Detailed Comparison of Continuations ----------

\end{text-sample}

Below is the rater model's response:

\begin{text-sample}
(placeholder since first line is not rendered for some reason)
**Fluency:** Both texts demonstrate good grammatical structure. Text A is a complete sentence, while Text B ends mid-sentence, but this isn't penalized as per the instructions.  Thus, they are roughly equal in fluency.

**Coherence:** Text A presents a seemingly disconnected statement.  Who is Palmer? Why are Austin and two boys being mentioned? Without further context, it lacks coherence. Text B describes Pearl's action and the scene she observes, creating a more coherent picture.

**Logicality:**  Text A lacks logicality due to the lack of context.  It's a grammatically correct sentence, but the information presented doesn't make much sense on its own. Text B describes a plausible scenario \u2013 someone running to a window and looking at a garden.

**Interestingness:** Text B is more interesting.  The imagery of the garden walk with closely shaven grass is more evocative than the seemingly random information presented in Text A.

Overall, Text B is superior due to its better coherence, logicality, and interestingness, despite both having good fluency.

My final verdict is: Text B is slightly better: [[B>A]]

\end{text-sample}

\section{Generated Samples}
\label{sec:further-generations}

\begin{figure*}[htbp]
	\begin{minipage}[b]{\linewidth}
		\centering		
            \centerline{\includegraphics[width=\linewidth]{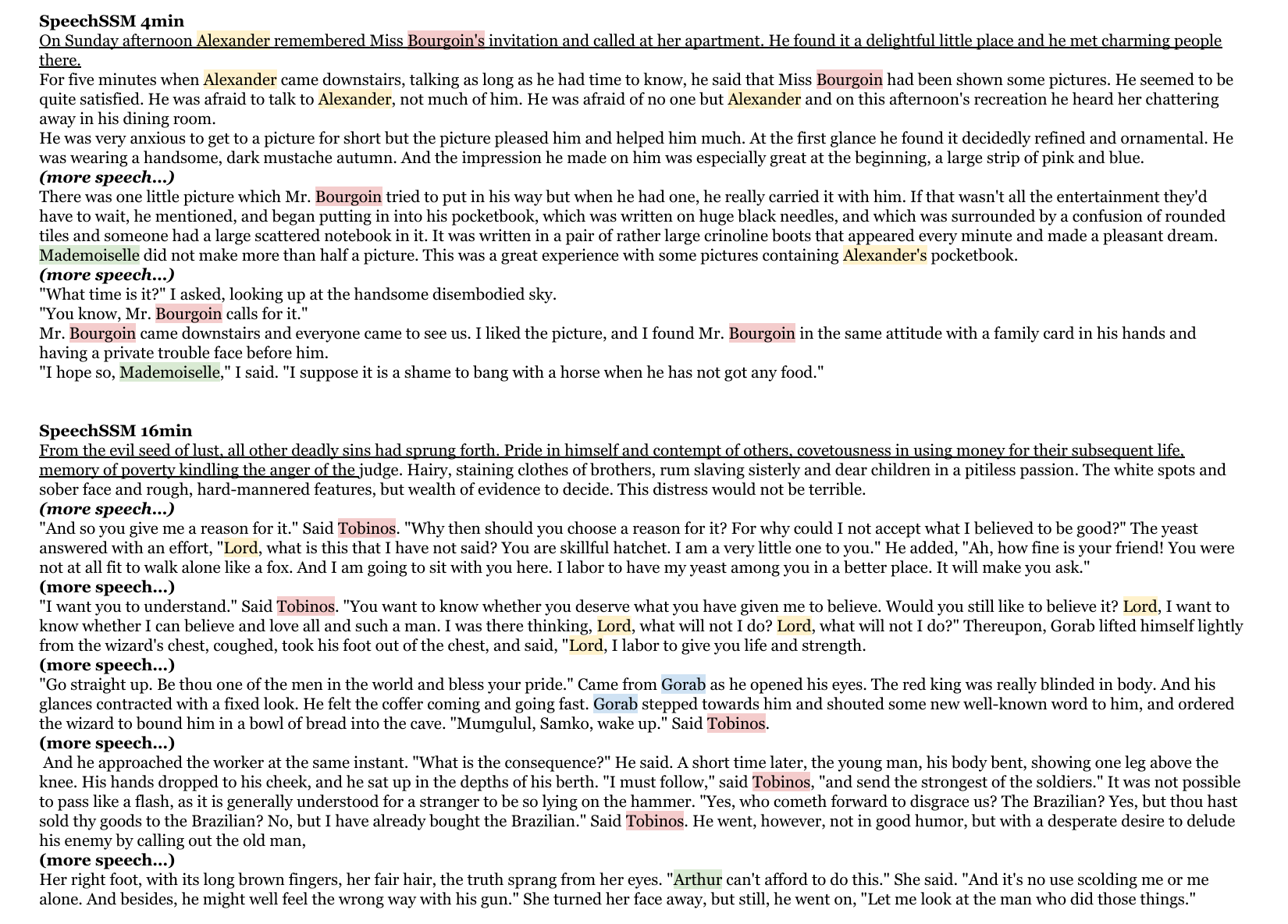}}
	\end{minipage}
    \vspace{-0.7cm}
	\caption{Automated transcripts of 4min and 16min speech continuations generated by SpeechSSM-2B, atop a 10-second audio-only prompt from our proposed LibriSpeech-Long (test-clean). The prompt is underlined, and parts abbreviated with \textit{(more speech...)} for emphasis. Recurring proper nouns are highlighted in consistent colors to show SpeechSSM's relative semantic consistency over time.}
  \vspace{-0.5cm}
  \label{fig:transcript_4min16min}
\end{figure*}

\begin{figure*}[t]
	\begin{minipage}[b]{\linewidth}
		\centering		
            \centerline{\includegraphics[width=\linewidth]{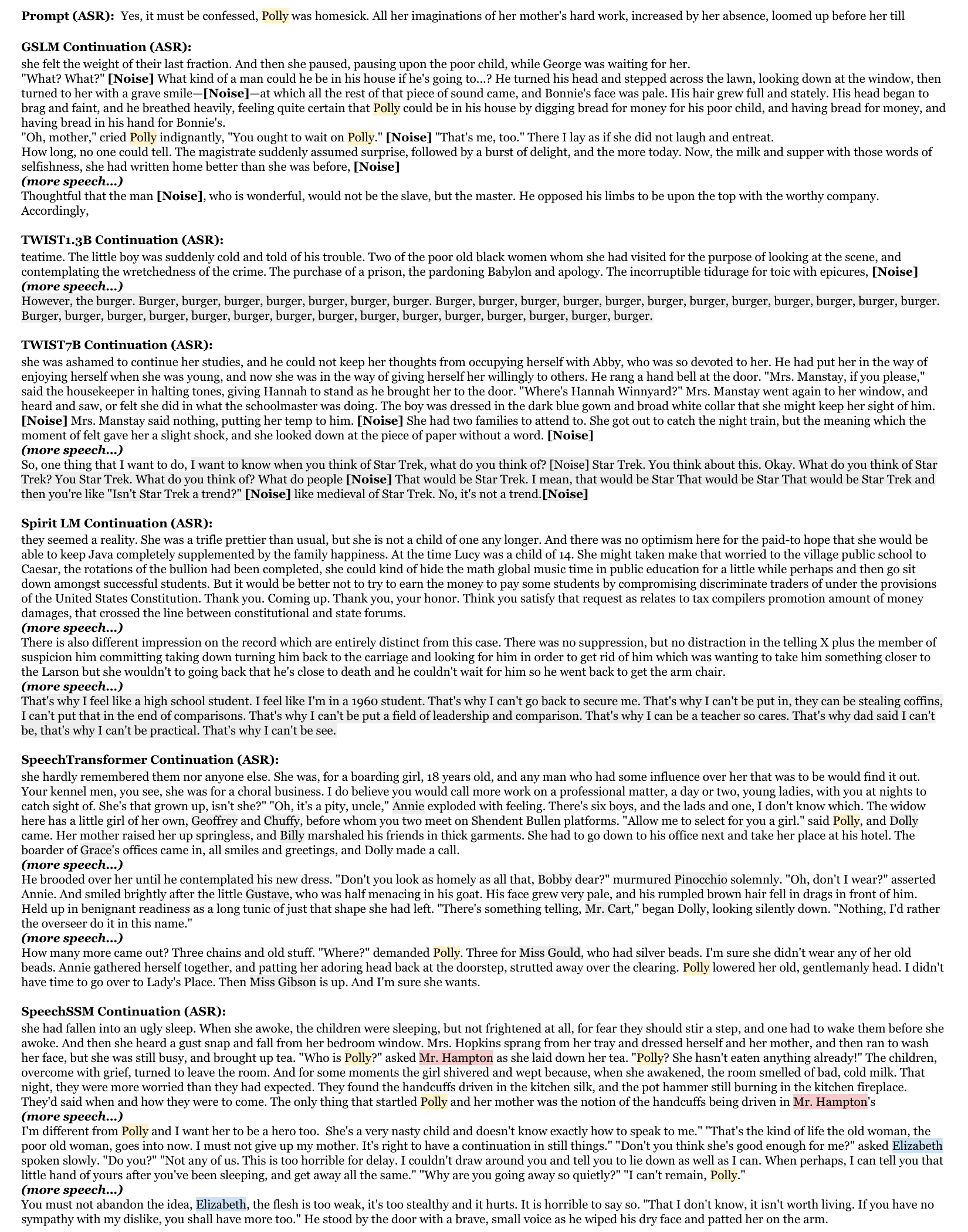}}
	\end{minipage}
    \vspace{-0.7cm}
	\caption{Automated transcripts of 4min speech continuations generated by SpeechSSM and baselines. Parts abbreviated with \textit{(more speech...)} for emphasis. Recurring proper nouns are highlighted in consistent colors to show SpeechSSM's relative semantic consistency over time. Non-sense sentences and proper noun errors are highlighted in grey.}
  \vspace{-0.5cm}
  \label{fig:transcript_comparison}
\end{figure*}

Figure \ref{fig:transcript_4min16min} presents transcripts of 4min and 16min speech continuations generated by SpeechSSM-2B, and Figure \ref{fig:transcript_comparison} compares 4min speech continuations generated by SpeechSSM-2B with those from other models. The recurring proper nouns are highlighted in consistent colors, demonstrating SpeechSSM's superior performance in semantic coherence and context maintenance over long generations. `4min' denotes the default SpeechSSM-2B and `16min' the variant trained on 16min segments.

\end{document}